\documentclass[10pt,twocolumn,letterpaper]{article}

\usepackage[pagenumbers]{cvpr} 

\usepackage{threeparttable}
\usepackage{enumerate}
\usepackage{graphicx}
\usepackage{makecell}
\usepackage{booktabs}
\usepackage{multirow}
\usepackage{multicol}
\usepackage{stfloats}
\usepackage{amsmath}
\usepackage{comment}
\usepackage{xcolor}
\usepackage{xspace}
\usepackage{array}

\allowdisplaybreaks
\definecolor{cvprblue}{rgb}{0.21,0.49,0.74}
\usepackage[pagebackref,breaklinks,colorlinks,allcolors=cvprblue]{hyperref}

\newcommand{\pname}{{Rascene}\xspace}

\def\paperID{6771} 
\def\confName{CVPR}
\def\confYear{2026}

\title{\!\!\pname: High-Fidelity 3D Scene Imaging with mmWave Communication Signals\!\!}

\author{Kunzhe Song \quad Geo Jie Zhou \quad Xiaoming Liu \quad Huacheng Zeng\\
Department of Computer Science and Engineering, Michigan State University\\
{\tt\small \{songkunz, geozhou, liuxm, hzeng\}@msu.edu}
}

\begin{document}

\twocolumn[{
\renewcommand\twocolumn[1][]{#1}
\maketitle\vspace{-0.2in}
\centering
\includegraphics[width=0.9\linewidth]{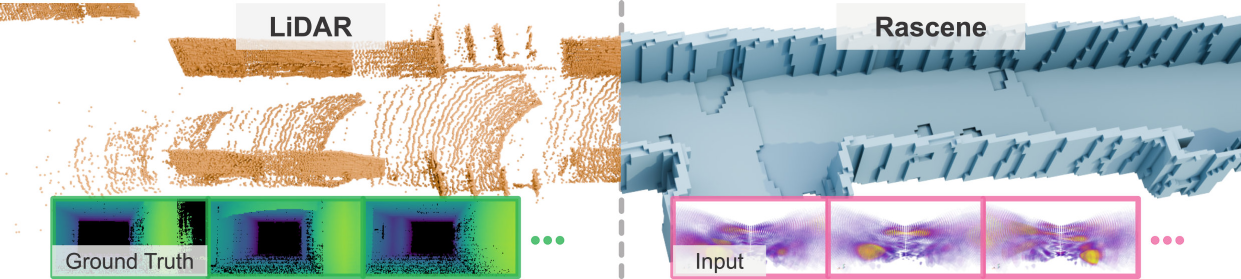}\vspace{-0.05in}
\captionof{figure}{High-fidelity 3D imaging generated by \pname from mmWave communication signals. 
We show that OFDM communication signals can support high-fidelity 3D imaging on a single device. 
Our multi-frame RF fusion suppresses multipath artifacts and integrates sparse observations into a complete 3D scene estimate (blue) that is closer to LiDAR ground truth (orange).}\vspace{0.2in}
\label{fig:teaser}
}]

\begin{abstract}

Robust 3D environmental perception is critical for applications such as autonomous driving and robot navigation. However, optical sensors such as cameras and LiDAR often fail under adverse conditions, including smoke, fog, and non-ideal lighting. Although specialized radar systems can operate in these environments, their reliance on bespoke hardware and licensed spectrum limits scalability and cost-effectiveness.
This paper introduces \pname, an integrated sensing and communication (ISAC) framework that leverages ubiquitous mmWave OFDM communication signals for 3D scene imaging. To overcome the sparse and multipath-ambiguous nature of individual radio frames, \pname performs multi-frame, spatially adaptive fusion with confidence-weighted forward projection, enabling the recovery of geometric consensus across arbitrary poses. Experimental results demonstrate that our method reconstructs 3D scenes with high precision, offering a new pathway toward low-cost, scalable, and robust 3D perception.


\end{abstract}
\vspace{-0.15in}

\section{Introduction}
\label{sec:intro}

Robust 3D environmental perception is critical for autonomous navigation and robotics. 
Existing 3D perception systems predominantly rely on cameras and LiDAR. 
However, camera-based methods \cite{Li_2025_CVPR, engel2017direct, mildenhall2021nerf, kerbl20233d} are fundamentally constrained by non-ideal lighting and fail in the presence of visual obscurants like smoke, fog, and snow. 
LiDAR systems, while offering accurate geometric measurements \cite{hess2016real, chen2019suma++, Chen2020OverlapNetLC}, remain expensive, bulky, and are similarly susceptible to adverse weather and occluding materials.

Radar-based 3D imaging has emerged as a compelling alternative \cite{lai2024enabling, 9823311, huang2025towards, guan2020through}, as it is robust to lighting variations and can sense through occlusions such as smoke and fog. 
However, practical deployment of \textit{specialized} radar systems remains challenging. 
These systems often require ultra-wideband (multi-GHz) spectrum allocations \cite{10161429, zhang2024towards}, which demand dedicated licenses or risk interference to incumbent systems. 
Furthermore, integrating bespoke sensing hardware increases cost, size, and power consumption \cite{adib2015capturing, abari2017enabling}, limiting suitability for compact, energy-constrained platforms such as AR/VR headsets and home robots.

This paper introduces \pname, an integrated sensing and communication (ISAC) framework that bridges this gap by utilizing millimeter-wave (mmWave) communication signals (\textit{e.g.}, 5G and Wi-Fi) for high-fidelity 3D imaging (Fig.~\ref{fig:teaser}). 
Tab.~\ref{tab:isac_comparison} provides a comparison between \pname and conventional sensing modalities. 
Unlike \textit{specialized} radar, \pname integrates sensing into communication systems. 
It leverages OFDM communication waveforms to extract fine-grained range and angle information without requiring dedicated sensing hardware or licensed spectrum. 
By leveraging communication network infrastructure, \pname provides a scalable and low-cost pathway for robust 3D perception on commodity Wi-Fi and cellular devices.

The design of \pname overcomes two fundamental challenges. 
The first is reliable RF data acquisition. 
Most communication-based sensing systems are \textit{bistatic}, relying on separate transmitter (Tx) and receiver (Rx). 
The dynamic relative poses of Tx and Rx devices introduce fundamental uncertainty. 
To address this, \pname is built on our key finding (Sec.~\ref{sec:ofdm-signal}) that commodity mmWave devices can operate in full-duplex mode for \textit{monostatic} sensing. 
The high directionality of phased-array antennas and short carrier wavelength provide sufficient Tx/Rx isolation. 
This capability allows \pname to perform precise Channel Impulse Response (CIR) measurements using co-located antennas, eliminating pose uncertainties and enabling the generation of 3D point clouds analogous to those from FMCW radar.

The second challenge stems from the inherent nature of RF signals. Single-frame RF observations are low-resolution, sparse, and corrupted by multipath. To address this, we introduce a multi-frame 3D imaging network that fuses arbitrarily posed observations, where each frame contributes a latent feature volume with confidence-aware forward projection. 
This source-driven fusion enforces geometric consensus, suppresses multipath-induced ghost or hallucinated structures, and improves scene completeness.

We collect a large-scale dataset across diverse indoor environments to evaluate \pname. Experiments demonstrate \pname's strong cross-scene generalization. Under within-dataset evaluation, \pname achieves the best overall performance among all baselines. Furthermore, \pname benefits substantially from multi-frame inputs, validating the effectiveness of the proposed multi-frame fusion mechanism.


\begin{table}[t]
\centering
\caption{Comparison of \pname and conventional modalities.
[Keys: O.P.=\textit{Obstacle Penetration}, O.R.=\textit{Occlusion Resilience}, S.L.E.=\textit{Spectrum License Exempt}, P.C.=\textit{Power Consumption}.]}
\vspace{-0.05in}
\resizebox{\linewidth}{!}{
\begin{tabular}{lccccccccc}
\toprule
\!\!\!\!\textbf{Tech.} & \!\!\!\!\textbf{Medium} & \!\!\!\textbf{Waveform} & \!\!\!\textbf{O.P.} & \!\!\!\textbf{O.R.} & \!\!\!\textbf{S.L.E.} & \!\!\!\textbf{Hardware} & \!\!\textbf{P.C.} & \!\!\!\textbf{Cost} & \!\!\!\textbf{Scalable} \\
\midrule
\!\!\!\!Camera & \!\!\!\!Light & \!\!\!--- & \!\!\!No & \!\!\!Poor & \!\!\!--- & \!\!\!Dedicated & \!\!Low & \!\!\!Low & \!\!\!High \\
\!\!\!\!LiDAR & \!\!\!\!Laser & \!\!\!Pulsed & \!\!\!No & \!\!\!Poor & \!\!\!--- & \!\!\!Dedicated & \!\!High & \!\!\!High & \!\!\!Low \\
\!\!\!\!Radar\!\! & \!\!\!\!Radio & \!\!\!FMCW, etc. & \!\!\!Yes & \!\!\!Good & \!\!\!No & \!\!\!Dedicated & \!\!Med & \!\!\!Med & \!\!\!Low \\
\!\!\!\!\pname & \!\!\!\!Radio & \!\!\!OFDM & \!\!\!Yes & \!\!\!Good & \!\!\!Yes & \!\!\!Reused & \!\!Low & \!\!\!$\sim$Zero & \!\!\!High \\
\bottomrule
\end{tabular}
}\vspace{-0.1in}
\label{tab:isac_comparison}
\end{table}

The main contributions of this work are as follows:
\begin{itemize}
\item We introduce \pname, an ISAC framework that enables 3D imaging using mmWave OFDM communication signals, without dedicated sensing hardware or spectrum.

\item We propose a multi-frame, confidence-aware fusion framework for RF signals to overcome the inherent challenges of noise, sparsity, and multipath corruption.

\item Experiments on our collected dataset show that \pname reconstructs scenes with high fidelity and outperforms baselines in within-dataset evaluation.
\end{itemize}

\section{Related Work}
\label{sec:related-work}

\subsection{RF Sensing}

Prior RF sensing systems mainly relied on specialized radar hardware, with FMCW radar being a prevalent technology \cite{adib2015capturing, adib20143d, adib2015multi, zhao2016emotion, abari2017enabling}. 
However, the deployment of these solutions is constrained by their reliance on bespoke, ultra-wideband hardware. 
For instance, some radar imaging systems utilize bandwidths up to 4\,GHz \cite{10161429, zhang2024towards, 9823311, lai2024enabling}, which require dedicated spectrum licenses from FCC. 

Recently, ISAC \cite{dong2022sensing, dilling2014isac, cheng2022integrated, 10158711, song2026spectrum, keskin2025fundamental, wu2024sensing, meneghello2023toward} has emerged as a promising paradigm, which reuses existing communication signals and infrastructures for sensing.
By exploiting Channel State Information (CSI) characteristics, sub-6\,GHz Wi-Fi networks have been utilized for a variety of sensing tasks \cite{kotaru2015spotfi, soltanaghaei2018multipath, Yan_2024_CVPR, song2024siwis}.
While attractive for deployment, such systems are constrained by their bistatic setup, narrow bandwidth, and small antenna arrays, which makes high-fidelity 3D imaging challenging.
In contrast, \pname performs monostatic sensing in the mmWave band and introduces adaptive multi-frame fusion for 3D imaging.

\subsection{Multi-View 3D Imaging}

\noindent\textbf{Imaging from Dense Visual Data.}
Multi-view 3D imaging is a central topic in computer vision. 
Classical methods \cite{campos2021orb, engel2014lsd, engel2017direct, newcombe2011dtam} exploit photometric consistency across calibrated images to estimate depth maps. 
More recently, neural implicit representations \cite{mildenhall2021nerf, kerbl20233d, Wu_2024_CVPR, Li_2023_CVPR} have achieved strong performance in novel view synthesis and scene reconstruction. 
A key assumption of these methods is the availability of dense and informative visual observations, where geometry can be inferred through reliable feature correspondence or photometric optimization across pixels.

\vspace{0.02in}
\noindent\textbf{Imaging from Sparse LiDAR Data.}
More closely related to our setting are methods that reconstruct scenes from sparse point clouds, typically captured by LiDAR sensors.
While sparse, LiDAR directly measures scene geometry with high fidelity. 
Hence, aggregating multiple LiDAR frames, often registered using algorithms like ICP or SLAM \cite{hess2016real, pan2021mulls, chen2019suma++, droeschel2018efficient}, is relatively straightforward. 
The fused point clouds are often dense enough to be directly processed \cite{rist2021semantic, Xia2023SCPNetSS, hou2022point, Zhu2020CylindricalAA} or converted into other representations like TSDF \cite{chibane2020implicit, chibane2020neural, mescheder2019occupancy, park2019deepsdf, chen2019learning, genova2019learning} for surface reconstruction. 

\section{RF Data Acquisition and Representation}
\label{sec:ofdm-signal}

\begin{figure}
    \centering
    \includegraphics[width=\linewidth]{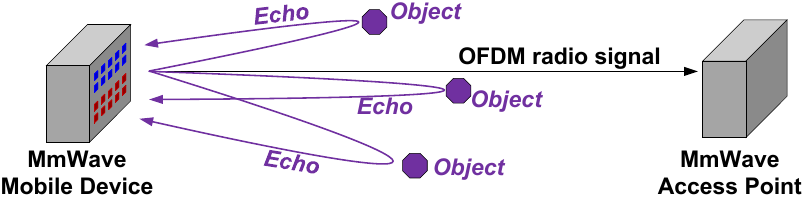}\vspace{-0.05in}
    \caption{Illustration of monostatic sensing in a mmWave communication system. The mmWave device (left) simultaneously transmits and receives OFDM communication signal  for  sensing.}\vspace{-0.15in}
    \label{fig:system_overview}
\end{figure}

MmWave communications have been widely adopted in both 5G and Wi-Fi standards to meet the ever-increasing demand for high data rates and low-latency connectivity. 
Consider a mobile mmWave communication device as shown in Fig.~\ref{fig:system_overview}. 
When it transmits data packets, the emitted OFDM RF signals inherently illuminate the surrounding objects. 
If the device is capable of transmitting and receiving simultaneously, it can capture the backscattered signals and estimate the Channel Impulse Response (CIR), thereby enabling precise object ranging without requiring dedicated sensing hardware, sensing waveform, or spectrum license. 
In practice, mmWave communication devices can indeed operate in full-duplex mode for \textit{monostatic} sensing thanks to their highly directional phased-array antennas and short carrier wavelength, which together provide sufficient RF isolation between transmission and reception paths \cite{wu2020mmtrack}.

\subsection{CIR-based Ranging}

Consider an OFDM symbol embedded in a 5G or Wi-Fi data packet. 
Denote $X(k) \in \mathbb{C}$ as the data on OFDM subcarrier $k$, where $k = 0, 1, \dots, K\!\!-\!\!1$ and $K$ is the total number of subcarriers. 
Since the device is transmitting and receiving simultaneously, it receives a reflected copy of its transmitted OFDM signal from surrounding objects.
Let $Y(k) \in \mathbb{C}$ denote the received data on subcarrier $k$.
The channel response on subcarrier $k$ can be estimated as $\hat{H}(k) = \frac{Y(k)}{X(k)}$.
Then, the corresponding CIR can be computed by:
$
\mathbf{\hat{h}} \triangleq
[\hat{h}(0), \hat{h}(1), \dots, \hat{h}(K\!\!-\!\!1)] = \operatorname{IFFT}\big([\hat{H}(0), \hat{H}(1), \ldots, \hat{H}(K\!-\!1)]\big)
$,
where $\hat{h}(n) \in \mathbb{C}$ denotes the coefficient of the $n$-th tap of the time-domain channel, $n = 0, 1, \dots, K\!\!-\!\!1$. 

Since the transmitter and receiver are co-located on the same device, their clock frequencies and data flow timing are synchronized. 
This synchronization allows the estimated CIR to be directly used for object ranging. 
Specifically, a strong magnitude $|\hat{h}(n)|$ indicates the presence of an object at a distance of $\frac{nc}{2B}$, where $c$ is the light speed and $B$ is the OFDM signal bandwidth. 
This capability fundamentally distinguishes the \textit{monostatic} configuration adopted in \pname from conventional \textit{bistatic} systems, where the transmitter and receiver are physically separated and the estimated CIR cannot be used for object ranging.

\begin{figure}
    \centering
    \includegraphics[width=0.55\linewidth]{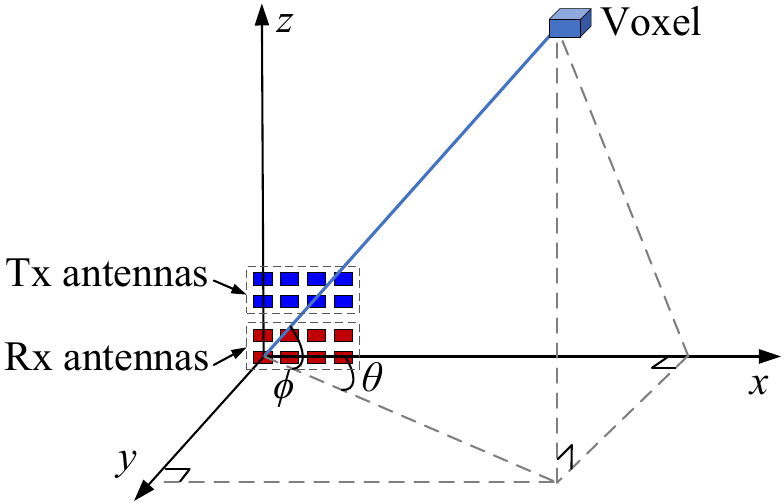}\vspace{-0.05in}
    \caption{Illustration of angular estimation on a mmWave device.}\vspace{-0.1in}
    \label{fig:ant_imaging}
\end{figure}

\subsection{Radio Imaging via Spatial Projection}

\begin{figure}
    \centering
    \begin{tabular}{>{\centering\arraybackslash}m{0.28\linewidth}
    >{\centering\arraybackslash}m{0.28\linewidth}
    >{\centering\arraybackslash}m{0.28\linewidth}}
            \textbf{Image} & $\mathbf{S}_{>0.1}$ &  $\mathbf{S}_{>0.2}$  \\
            \includegraphics[width=\linewidth]{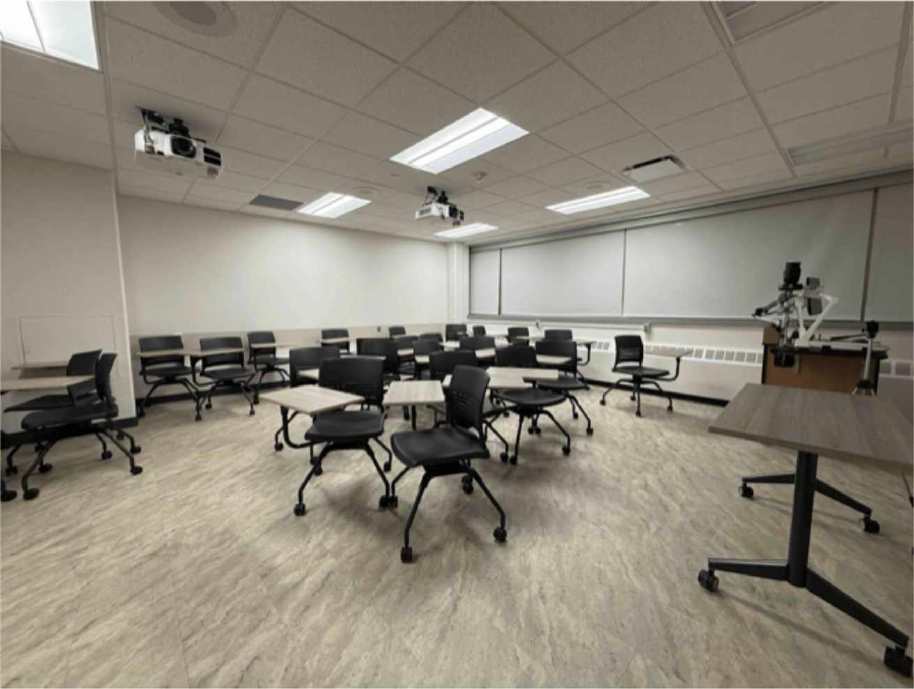} &
            \includegraphics[width=\linewidth]{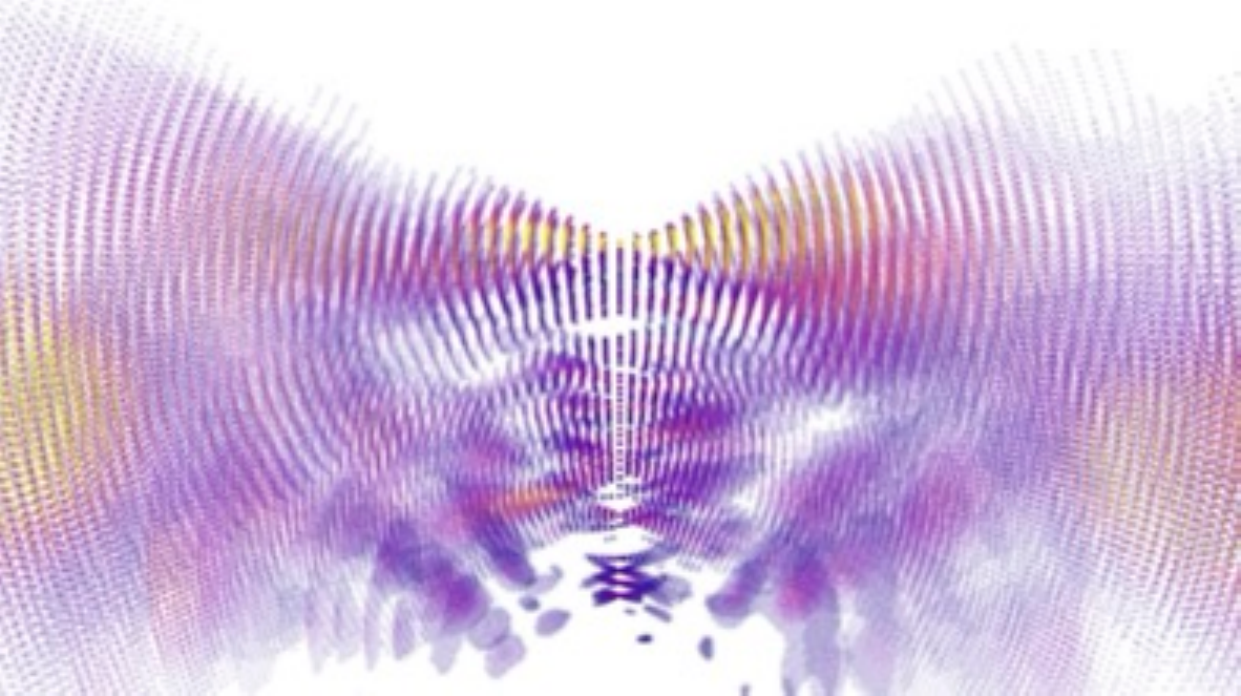} &
            \includegraphics[width=\linewidth]{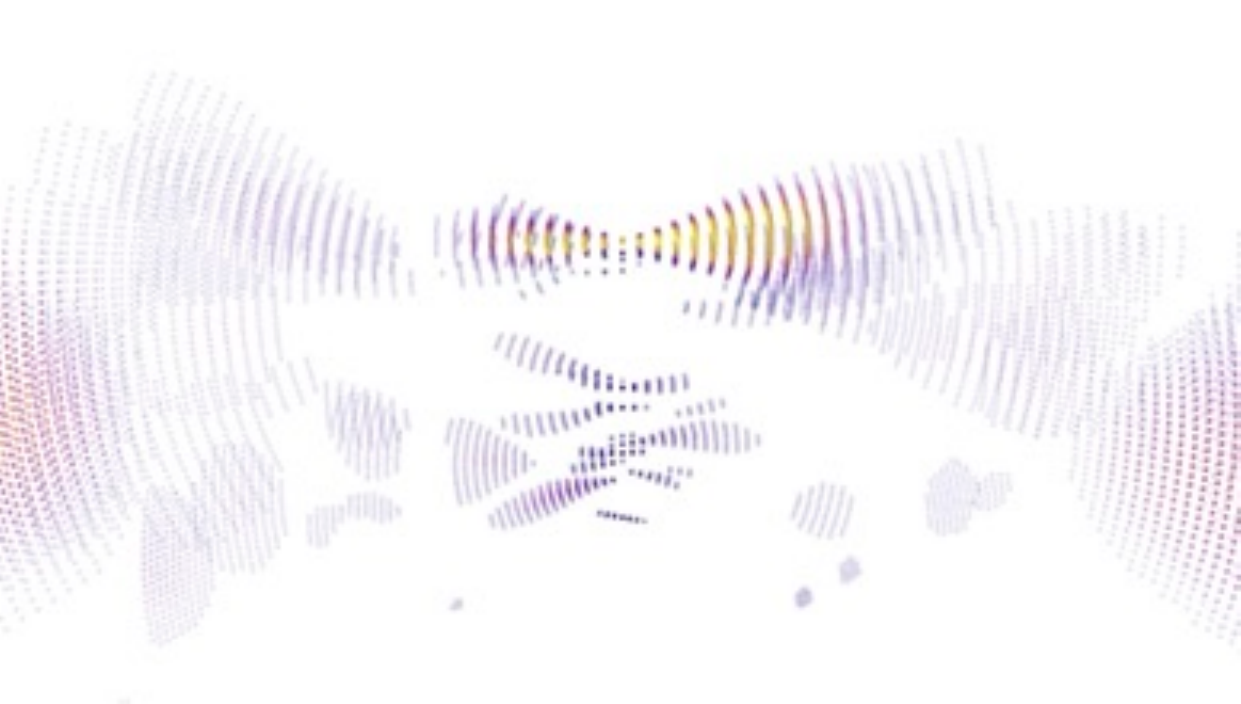}  \\
            
            \includegraphics[width=\linewidth]{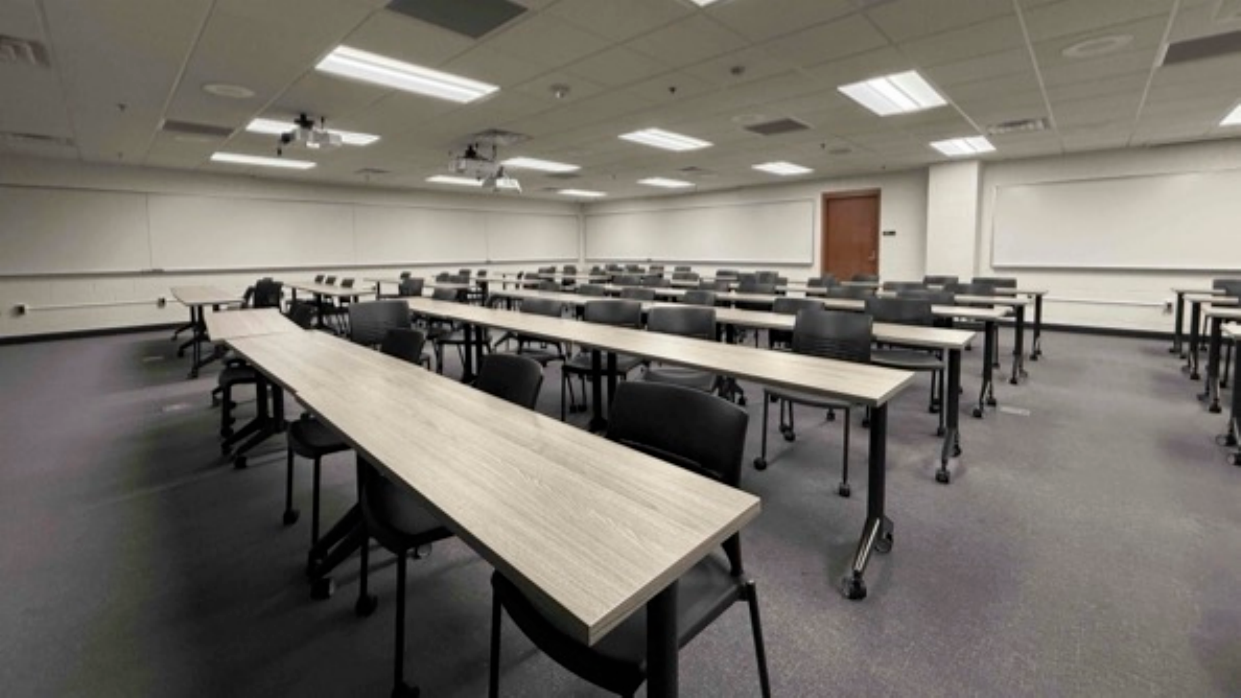} &
            \includegraphics[width=\linewidth]{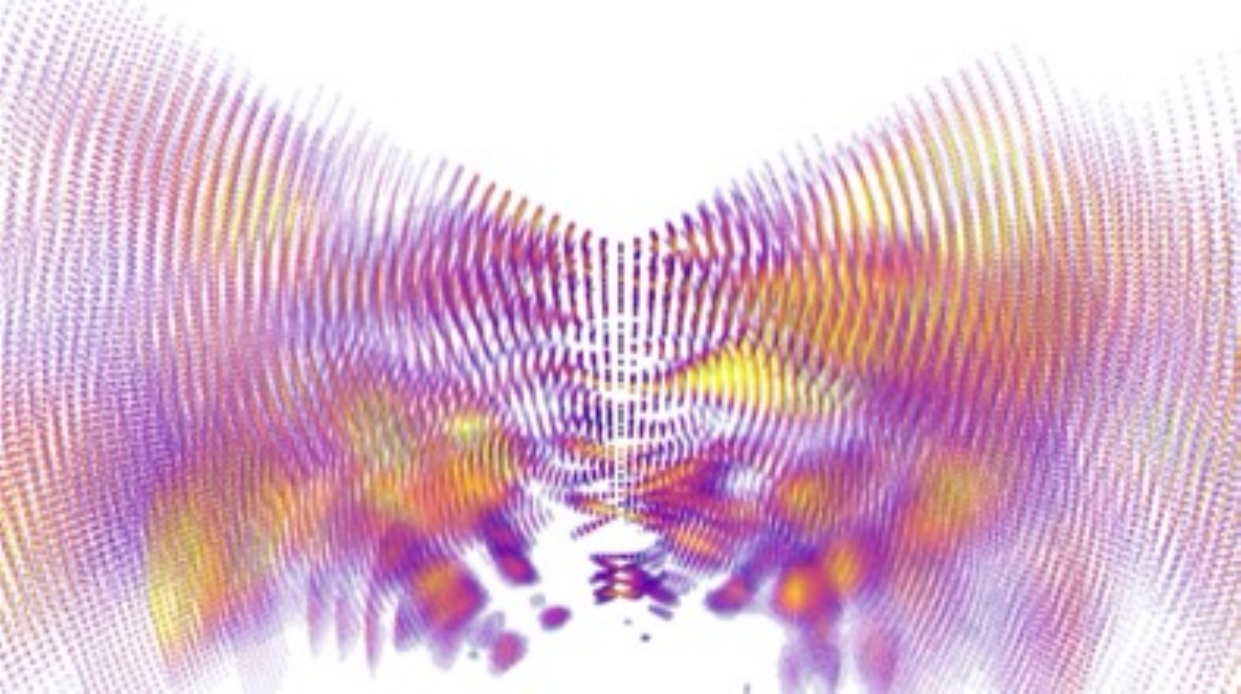} &
            \includegraphics[width=\linewidth]{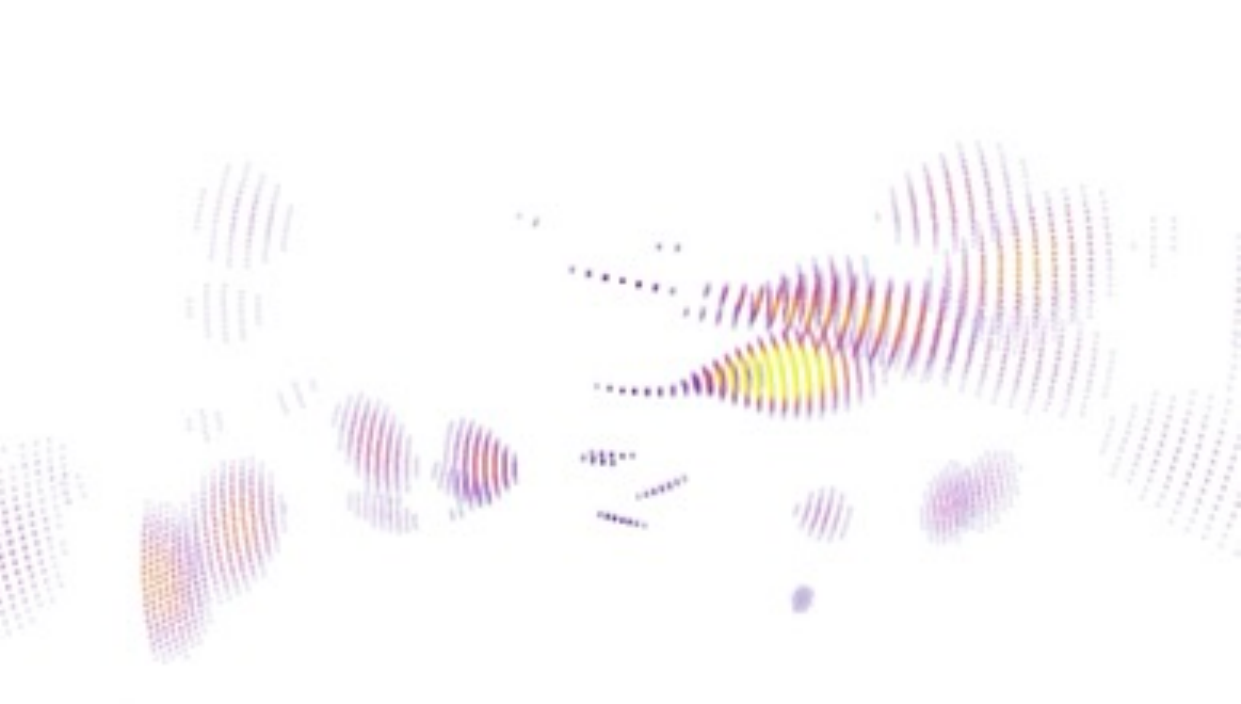} \\
            
            \includegraphics[width=\linewidth]{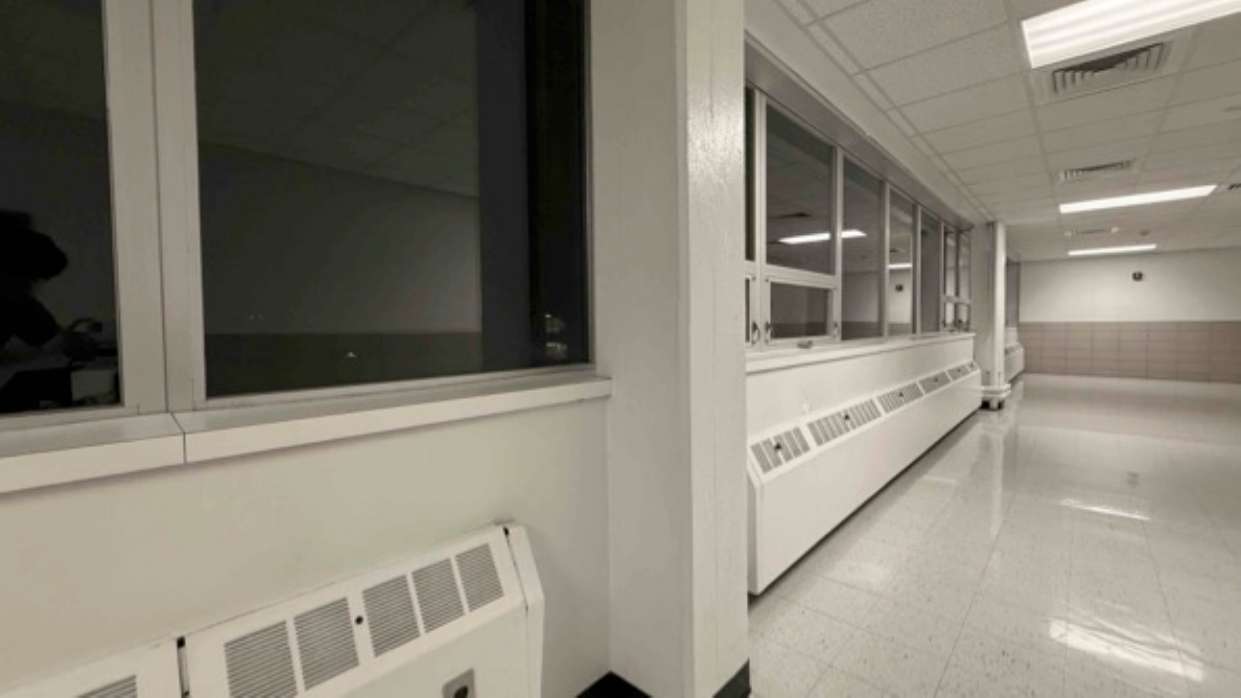} &
            \includegraphics[width=\linewidth]{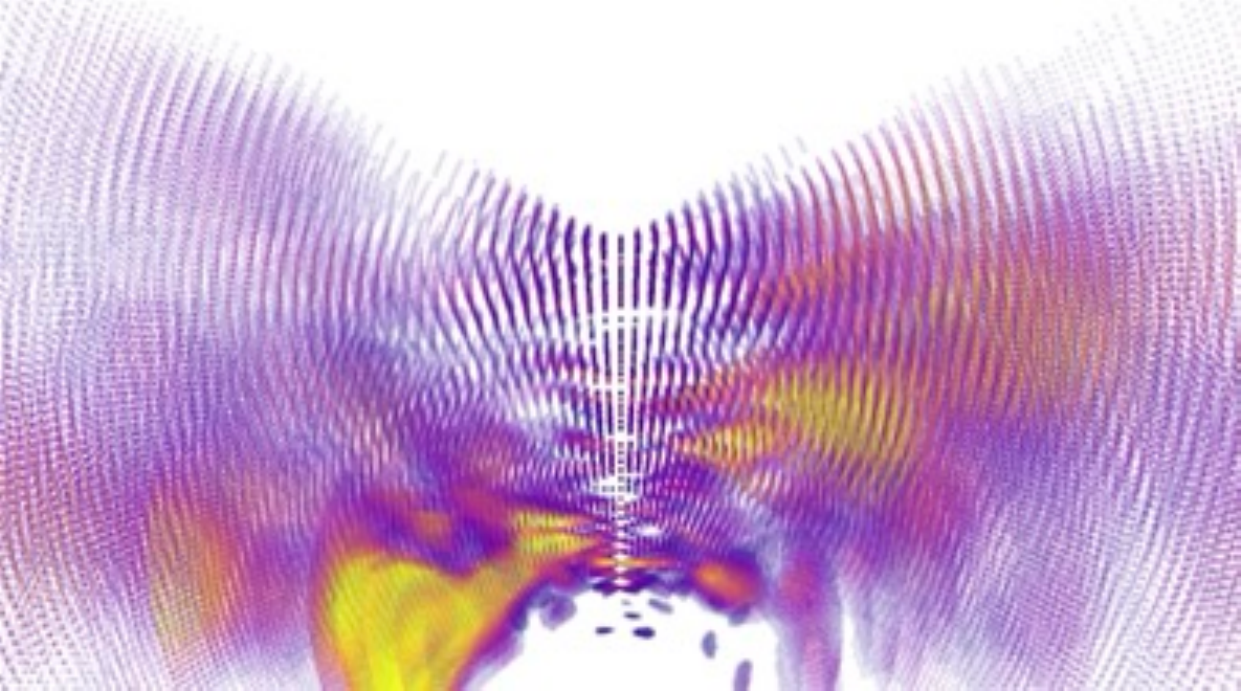} &
            \includegraphics[width=\linewidth]{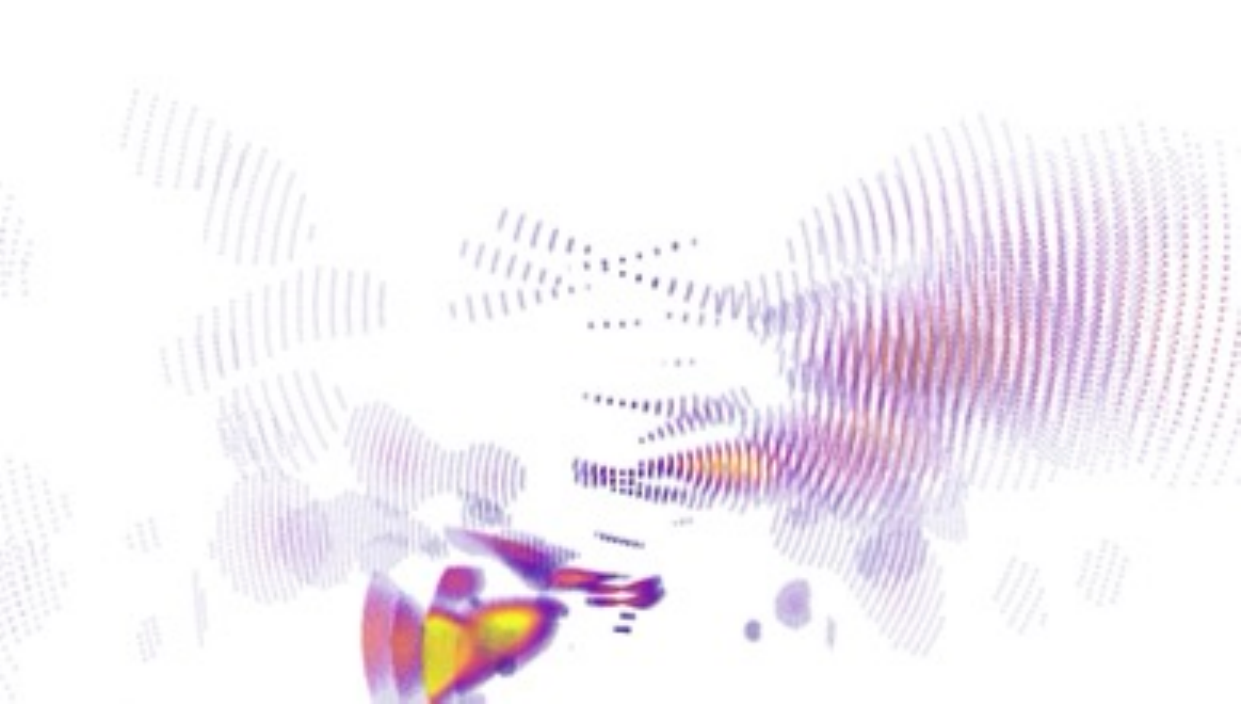} \\
            \end{tabular}\vspace{-0.05in}
    \caption{Examples of generated radio point clouds, where     $\mathbf{S}_{>a} = \{s \!\!\in\!\! \mathbf{S} \, | \, s \!\!>\!\! a \}$. Threshold $a$ is set to 0.1 and 0.2, and the 3D point clouds are projected onto 2D images for for ease of visualization.}\vspace{-0.10in}
    \label{fig:radio_frame}
\end{figure}

Angular estimation is another critical component of radio imaging, as resolving the angular positions of reflections enables objects to be separated spatially. 
Commercial mmWave communication devices are widely equipped with phased-array antennas, which can be leveraged for angular estimation without requiring additional sensing hardware.

Referring to Fig.~\ref{fig:ant_imaging}, we define the patch antenna panel as the $x$-$z$ plane, \textit{i.e.}, the plane $y=0$. 
Consider a voxel in the 3D space, and let $\theta$ and $\phi$ be its azimuth and elevation angles, respectively.
The unit vector pointing from the antenna array to this voxel can then be written as:
\begin{equation}
\vec{u}(\theta, \phi) = 
[\cos(\theta)\cos(\phi), 
~\cos(\theta)\sin(\phi), 
~\sin(\phi)].  
\end{equation}

Let $\mathcal{T}$ and $\mathcal{R}$ denote the sets of Tx and Rx antennas, respectively. 
Denote $\vec{p}_i = (x_i, 0, z_i)$ as the coordinate of Tx antenna $i \in \mathcal{T}$ and 
$\vec{q}_j = (x_j', 0, z_j')$ as the coordinate of Rx antenna $j \in \mathcal{R}$. 
When antenna $i$ transmits radio signal and antenna $j$ receives the backscattered signal from a voxel in direction $(\theta, \phi)$, the additional signal propagation distance relative to the antenna array center is 
$\langle\vec{u}(\theta, \phi), ~ \vec{p}_i + \vec{q}_j\rangle$, where $\langle \cdot, \cdot \rangle$ is inner product.
Thus, the complex weight associated with the antenna pair $(i, j)$ can be written as:
\begin{equation}
    w_{i,j}(\theta, \phi) = \exp\big(-j\frac{2\pi}{\lambda} 
    \langle\vec{u}(\theta, \phi), ~ \vec{p}_i + \vec{q}_j\rangle\big),
    \label{eq:weights}
\end{equation}
where $\lambda$ is the wavelength of carrier radio signal.

Let 
$\mathbf{\hat{h}}_{ij} = 
[\hat{h}_{ij}(0), \hat{h}_{ij}(1), \dots, \hat{h}_{ij}(K\!-\!1)]$
be the CIR measurement when \pname uses antenna pair $(i,j)$ for signal transmission and reception.
Then, the reflected signal strength of voxel $(n, \theta, \phi)$ can be computed by:
\begin{equation}
    s(n, \theta, \phi) 
    = 
    \left|\sum_{i \in \mathcal{T}}  \sum_{j \in \mathcal{R}} w_{i,j}(\theta, \phi) \hat{h}_{ij}(n)\right|,  
    \label{eq:beamformed_csi}
\end{equation}
where $n$ is the voxel distance index, with absolute distance being $r = \frac{nc}{2B}$.
A large value of $s(n, \theta, \phi)$ indicates the presence of an object, and a small value indicates its absence. 

Based on Eq.~\eqref{eq:beamformed_csi}, a radio frame, represented by the point clouds in the spherical coordinate system, can be written as: 
\begin{equation}
    \mathbf{S} = \left\{s(n, \theta, \phi) \; | \;\; n \in \mathbf{N}, \theta \in \mathbf{\Theta}, \phi \in \mathbf{\Phi} \right\}.
    \label{eq:radio_frame}
\end{equation}

Fig.~\ref{fig:radio_frame} shows three radio frames of point clouds generated by a mmWave communication device. 

\begin{figure*}
    \centering
    \includegraphics[width=0.97\textwidth]{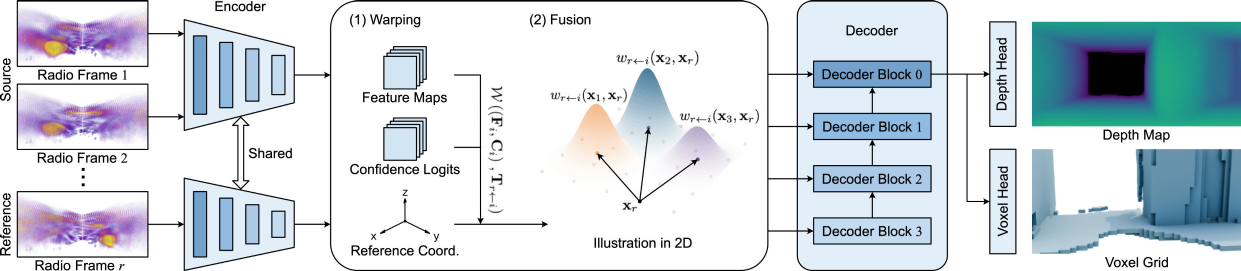}\vspace{-0.05in}
    \caption{Overview of the multi-frame 3D RF imaging network. 
    Given multiple radio frames and poses, a shared encoder predicts per-frame feature maps and confidence logits. 
    We then warp all features to a reference frame and fuse them into a unified representation. 
    A coarse-to-fine 3D decoder with voxel and depth heads outputs the reconstructed voxel grid and depth map.}\vspace{-0.1in}
    \label{fig:network-overview}
\end{figure*}

\section{Multi-Frame 3D RF Imaging}
\label{sec:imaging}

Given radio frames
\(\mathcal{S} = \{ \mathbf{S}_i \in \mathbb{R}^{|\mathbf{N}| \times |\mathbf{\Theta}| \times |\mathbf{\Phi}|}\}_{i=1}^N\)
captured at known poses
\(\mathcal{G}=\{\mathbf{G}_i \in SE(3)\}_{i=1}^{N}\),
our objective is to learn a mapping function \(\mathcal{F}\) that estimates a sequence of dense 3D voxel grids
\(\mathcal{V}=\{\hat{\mathbf{V}}_i \in \mathbb{R}^{X \times Y \times Z}\}_{i=1}^{N}\)
and corresponding depth maps
\(\mathcal{D}=\{\hat{\mathbf{D}}_i \in \mathbb{R}^{H \times W}\}_{i=1}^{N}\)
from the radio frames captured along arbitrary motion trajectories:
\begin{align}
    (\mathcal{V}, \mathcal{D}) = \mathcal{F}(\mathcal{S}, \mathcal{G}).
\end{align}

The main challenge lies in cross-frame fusion under sparse and multipath-corrupted RF observations.
We address this with an RF imaging framework (Sec.~\ref{sec:imaging-framework}) and a spatially adaptive warping and fusion module (Sec.~\ref{sec:warping-fusion}) that enforces geometric consistency across arbitrary poses.
An overview of the proposed framework is shown in Fig.~\ref{fig:network-overview}.

\subsection{RF Imaging Framework}
\label{sec:imaging-framework}

Unlike conventional multi-view reconstruction, RF observations are neither texture-rich like RGB images nor geometrically explicit like LiDAR points.
Instead, radio frames are strongly affected by attenuation and multipath (Fig.~\ref{fig:radio_frame}), which makes single-frame 3D imaging ill-posed.

Following Sec.~\ref{sec:ofdm-signal}, each \(\mathbf{S}_i\) is transformed from spherical measurement space \((n,\theta,\phi)\) into a Cartesian volume in the local sensor frame.
For simplicity, we reuse \(\mathbf{S}_i\) to denote this Cartesian representation.
We then choose a reference frame \(r \in \{1,\dots,N\}\) and transform all frames into its coordinate system.
A shared encoder \(\mathcal{E}\), parameterized by \(\boldsymbol{\theta}_{\mathcal{E}}\), maps each frame to a latent feature volume \(\mathbf{F}_i \in \mathbb{R}^{X\times Y\times Z\times C_F}\) and confidence logits \(\mathbf{C}_i \in \mathbb{R}^{X\times Y\times Z}\):
\begin{align}
    (\mathbf{F}_i, \mathbf{C}_i) = \mathcal{E}(\mathbf{S}_i \,; \boldsymbol{\theta}_{\mathcal{E}}).
\end{align}

Since \((\mathbf{F}_i,\mathbf{C}_i)\) are defined in local coordinates, we warp them into the reference frame \(r\) using the relative rigid transformation \(\mathbf{T}_{r\leftarrow i}\) and a warping operator \(\mathcal{W}(\cdot\,, \mathbf{T})\):
\begin{gather}
    \mathbf{T}_{r\leftarrow i} = \mathbf{G}_r^{-1}\mathbf{G}_i
    = \left[ \begin{array}{cc}
    \mathbf{R}_{r\leftarrow i} & \mathbf{t}_{r\leftarrow i} \\
    \mathbf{0}^{\top} & 1
    \end{array} \right], \\
    (\tilde{\mathbf{F}}_i, \tilde{\mathbf{C}}_i)
    = \mathcal{W}\left((\mathbf{F}_i, \mathbf{C}_i)\,, \mathbf{T}_{r\leftarrow i}\right).
    \label{eq:warping-operation}
\end{gather}

After warping, the fusion operator \(\mathcal{A}\) aggregates all warped features into a unified latent representation \(\mathbf{Z}_r\),
which captures geometric consensus across frames while suppressing multipath artifacts.
Because this fused representation is still sparse, we employ a coarse-to-fine 3D decoder \(\mathcal{O}\) to progressively densify \(\mathbf{Z}_r\) into a dense feature volume \(\mathbf{H}_r\).
Finally, two task-specific heads, \(\mathcal{H}_v\) and \(\mathcal{H}_d\), predict the voxel grid \(\hat{\mathbf{V}}_r\) and depth map \(\hat{\mathbf{D}}_r\) from \(\mathbf{H}_r\):
\begin{gather}
    \mathbf{Z}_r = \mathcal{A}\left(\{(\tilde{\mathbf{F}}_i,\tilde{\mathbf{C}}_i)\}_{i=1}^{N}\right), \label{eq:fusion-mechanism} \\
    \mathbf{H}_r = \mathcal{O}(\mathbf{Z}_r \,; \boldsymbol{\theta}_{\mathcal{O}}), \\
    \hat{\mathbf{V}}_r = \mathcal{H}_v(\mathbf{H}_r \,; \boldsymbol{\theta}_v), \quad
    \hat{\mathbf{D}}_r = \mathcal{H}_d(\mathbf{H}_r \,; \boldsymbol{\theta}_d).
\end{gather}

The multi-frame 3D RF imaging framework, parameterized by
\( \boldsymbol{\Theta} = \{ \boldsymbol{\theta}_\mathcal{E}, \boldsymbol{\theta}_\mathcal{O}, \boldsymbol{\theta}_v, \boldsymbol{\theta}_d \} \),
is optimized end-to-end. For a window of \(N\) frames, each frame is used once as the reference, and losses are summed over all references:
\begin{align}
    \mathcal{L}
    = \sum_{r=1}^{N}\left(
    \lambda_{v}\mathcal{L}_{\text{voxel}}^{(r)}
    + \lambda_{d}\mathcal{L}_{\text{depth}}^{(r)}
    \right),
\end{align}
where \(\lambda_{v}\) and \(\lambda_{d}\) are scalar hyperparameters that balance the two tasks.
The voxel loss \(\mathcal{L}_{\text{voxel}}^{(r)}\) is the binary cross-entropy loss between the predicted grid \(\hat{\mathbf{V}}_r\) and its corresponding ground truth \(\mathbf{V}_r^*\) over the volume \(\Omega\):
\begin{align}
    \mathcal{L}_{\text{voxel}}^{(r)} = &- \frac{1}{|\Omega|}\sum_{\mathbf{x} \in \Omega} [ \mathbf{V}_r^*(\mathbf{x})\log(\hat{\mathbf{V}}_r(\mathbf{x})) \nonumber\\
    &+\; (1-\mathbf{V}_r^*(\mathbf{x}))\log(1-\hat{\mathbf{V}}_r(\mathbf{x})) ],
\end{align}
and the depth loss \(\mathcal{L}_{\text{depth}}^{(r)}\) is the L1 loss between the predicted depth map \(\hat{\mathbf{D}}_r\) and its ground truth \(\mathbf{D}_r^*\):
\begin{align}
    \mathcal{L}_{\text{depth}}^{(r)} = \frac{1}{HW} \sum_{(u,v)} \left| \hat{\mathbf{D}}_r(u,v) - \mathbf{D}_r^*(u,v) \right|.
\end{align}

\subsection{Spatially Adaptive Warping and Fusion}
\label{sec:warping-fusion}

Instantiating the warping \(\mathcal{W}\) in Eq.~\eqref{eq:warping-operation} and fusion \(\mathcal{A}\) in Eq.~\eqref{eq:fusion-mechanism} operators is the core challenge of our framework.
Unlike LiDAR points, which are geometrically explicit, RF observations provide only indirect and highly ambiguous measurements of scene geometry.
As a result, direct cross-frame warping and aggregation can easily accumulate inconsistent evidence and spurious reflections.

To address this issue, we formulate both \(\mathcal{W}\) and \(\mathcal{A}\) as sparse projection and aggregation operators.
Rather than querying values at target voxels, each source voxel \(\mathbf{x}_i \in \Omega_i\) is transformed into the reference frame and contributes only within a local support region in the target grid.
This strategy avoids repeated sampling of empty target locations and better preserves sparse yet informative RF responses that are critical for reconstruction.
Given frame \(i\), the continuous reference-frame coordinate of source voxel \(\mathbf{x}_i\) is:
\begin{align}
    \tilde{\mathbf{x}}_{r \leftarrow i} = \mathbf{T}_{r \leftarrow i}\mathbf{x}_i.
\end{align}

The source feature $\mathbf{F}_i(\mathbf{x}_i)$ is then distributed onto the target grid $\Omega_r$ by contributing to neighboring voxels $\mathbf{x} \in \mathcal{N}(\tilde{\mathbf{x}}_{r \leftarrow i})$ within a given radius $R$.
Since our transformations are purely rigid, without anisotropic scaling or shearing, we employ a simple but effective isotropic Gaussian kernel $K_\sigma$ to model geometric contribution, controlled by a parameter $\sigma$:
\begin{align}
    K_{\sigma}(\tilde{\mathbf{x}}_{r \leftarrow i}, \mathbf{x})
    = \exp\!\left(
    -\frac{\|\tilde{\mathbf{x}}_{r \leftarrow i}-\mathbf{x}\|^2}{2\sigma^2}
    \right).
\end{align}

The fusion operator \(\mathcal{A}\) leverages this projection by performing a spatially adaptive, confidence-weighted average.
We define an adaptive weight \(w_{r \leftarrow i}(\mathbf{x}_i, \mathbf{x}_r)\) for the contribution of source voxel \(\mathbf{x}_i\) in frame \(i\) to target voxel \(\mathbf{x}_r\) in frame \(r\).
This weight combines the geometric proximity kernel \(K_\sigma\) with predicted signal reliability at the source, derived from the confidence logits \(\mathbf{C}_i(\mathbf{x}_i)\):
\begin{gather}
    w_{r \leftarrow i}(\mathbf{x}_i, \mathbf{x}_r)
    = K_\sigma(\tilde{\mathbf{x}}_{r \leftarrow i}, \mathbf{x}_r)
    \cdot \left[\alpha\!\left(\mathbf{C}_i(\mathbf{x}_i)\right)\right]^\eta,
    \label{eq:adaptive-weight}
\end{gather}
where $\alpha(c) = \log(1 + e^c)$ maps confidence logits $c$ to a non-negative reliability score.
We raise this reliability score to the power $\eta$, a hyperparameter that controls the sharpness of confidence weighting.

The final unified representation $\mathbf{Z}_r$ is then computed as a normalized weighted average.
By weighting each contribution using the learned non-linear reliability, this formulation ensures that the aggregate consensus is dominated by high-confidence geometric signals, yielding stable and robust fusion:
\begin{align}
    \mathbf{Z}_r(\mathbf{x}_r)
    = \frac{
    \sum_{i=1}^{N}\sum_{\mathbf{x}_i \in \Omega_i}
    w_{r \leftarrow i}(\mathbf{x}_i,\mathbf{x}_r)\mathbf{F}_i(\mathbf{x}_i)}
    {\sum_{i=1}^{N}\sum_{\mathbf{x}_i \in \Omega_i}
    w_{r \leftarrow i}(\mathbf{x}_i,\mathbf{x}_r)+\varepsilon},
\end{align}
where \(w_{r \leftarrow i}(\mathbf{x}_i,\mathbf{x}_r)=0\) when
\(\mathbf{x}_r \notin \mathcal{N}(\tilde{\mathbf{x}}_{r \leftarrow i})\), and
\(\varepsilon>0\) ensures numerical stability in low-support regions.
\(\Omega_i\) denotes the grid of source voxels in frame \(i\).

\section{Implementation}

\begin{figure}
    \centering
    \includegraphics[width=\linewidth]{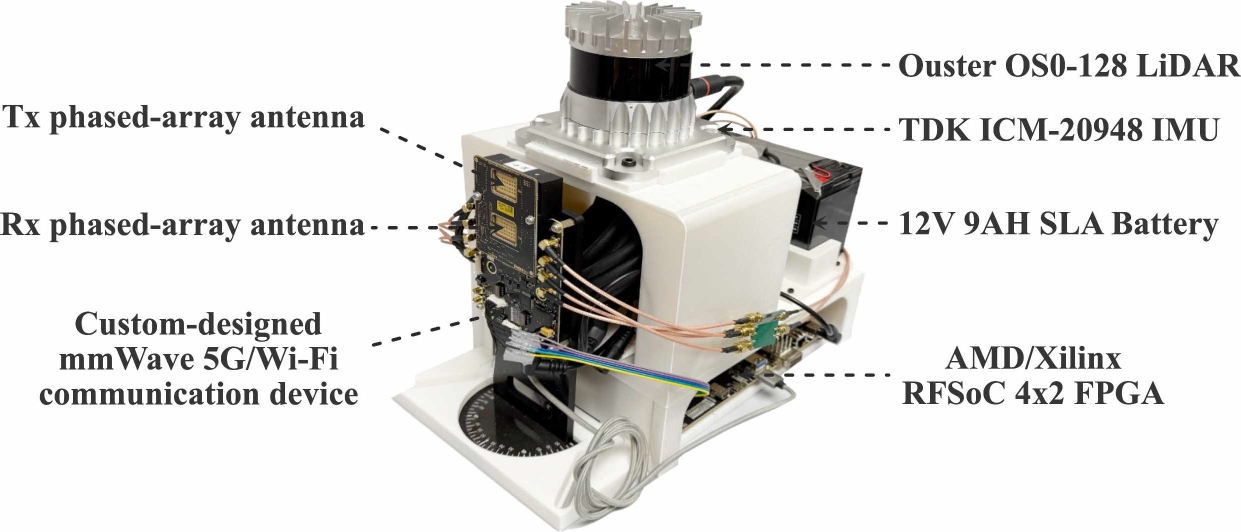}
    \caption{\pname data collection platform.}\vspace{-0.1in}
    \label{fig:fmcw_radar_testbed}
\end{figure}

\noindent\textbf{Hardware Setup.}
We built a prototype of \pname as shown in Fig.~\ref{fig:fmcw_radar_testbed} to evaluate its performance in realistic scenarios. 
The prototyped \pname comprises three main sensors: 
(a) a custom-designed mmWave 5G/Wi-Fi communication device that supports full-duplex operation for monostatic sensing, 
(b) a commercial LiDAR sensor, and 
(c) a commercial IMU. 
Our custom-designed communication device operates at 60 GHz with a bandwidth of 1.2288 GHz, featuring $16$ transmit (Tx) and $16$ receive (Rx) antenna elements. 
It provides an effective sensing range of 7 meters, covering a field of view (FoV) of 120$^\circ$ horizontally and 60$^\circ$ vertically. 
For ground truth, we use an Ouster 128-beam LiDAR to capture 3D point clouds. 
For 6-DoF pose acquisition, we integrate a TDK ICM-20948 9-axis IMU. 

\vspace{0.02in}
\noindent\textbf{Model Architecture.}
Each RF frame is projected to a Cartesian voxel grid of size $64\times64\times32$ (12\,cm voxel size).
The encoder $\mathcal{E}$ and decoder $\mathcal{O}$ both use four convolutional stages, with channel multipliers $(1,2,4,8)$ relative to the stem channels.
Following Sec.~\ref{sec:warping-fusion}, we perform stage-wise warping $\mathcal{W}$ and fusion $\mathcal{A}$ after each encoder stage to align and aggregate multi-view evidence in the reference frame.
For the sparse projection operator, we use neighborhood radius $R=2$ and confidence sharpness $\eta=3$.

\vspace{0.02in}
\noindent\textbf{Data Collection.}
We collected synchronized RF-LiDAR frame pairs across 20 indoor environments.
The LiDAR and ISAC streams are recorded at 10\,Hz while the platform moves at 0.5\,m/s.
We sample one frame every 2\,s and group five sampled frames into a window, 
which preserves sufficient overlap for geometric consistency while providing meaningful viewpoint diversity.

\begin{table*}[!t]
\footnotesize
\centering
\begin{threeparttable}
\caption{Quantitative evaluation results of \pname's cross-scene generalization performance. 
The model is trained on 12 scenarios and evaluated on 8 distinct, unseen test scenarios (A-H). 
We report metrics for both depth estimation (AbsRel, MAE, RMSE) and 3D voxel reconstruction (CD, CD$_{\text{Diag}}$). 
Lower values are better for all metrics.}\vspace{-0.05in}
\begin{tabular}{
m{0.06\textwidth}<{\centering}
m{0.08\textwidth}<{\centering}
m{0.06\textwidth}<{\centering}
m{0.06\textwidth}<{\centering} 
m{0.06\textwidth}<{\centering} 
m{0.06\textwidth}<{\centering} 
m{0.06\textwidth}<{\centering}
m{0.06\textwidth}<{\centering}
m{0.06\textwidth}<{\centering}
m{0.06\textwidth}<{\centering}
m{0.06\textwidth}<{\centering}}
\toprule
\multicolumn{2}{c}{Metric} & A & B & C & D & E & F & G & H & Avg. \\ 
\midrule
\multirow{3}{*}{\vspace{-0.01in} Depth} & AbsRel (\%) & 10.1 & 4.9 & 4.5 & 3.3 & 17.3 & 19.4 & 7.4 & 8.7 & 9.4 \\
& MAE (cm) & 22.9 & 11.5 & 10.2 & 7.3 & 36.2 & 37.4 & 16.8 & 19.9 & 20.2 \\
& RMSE (cm) & 43.2 & 21.2 & 22.2 & 17.7 & 63.3 & 66.9 & 34.4 & 38.1 & 38.0 \\
\midrule
\multirow{2}{*}{\vspace{-0.01in} Voxel} & CD  (cm) & 20.4 & 15.4 & 13.0 & 9.1 & 37.0 & 26.5 & 16.8 & 16.9 & 19.7 \\
& CD$_{\text{Diag}}$(\%) & 2.5 & 1.7 & 1.5 & 1.1 & 3.7 & 3.3 & 2.0 & 2.1 & 2.3 \\
\bottomrule
\end{tabular}
\label{table:quantitative-performance}
\end{threeparttable}
\end{table*}

\begin{figure*}[!t]
    \centering
    \begin{minipage}{0.95\textwidth}\vspace{0.1in}
    \includegraphics[width=\textwidth]{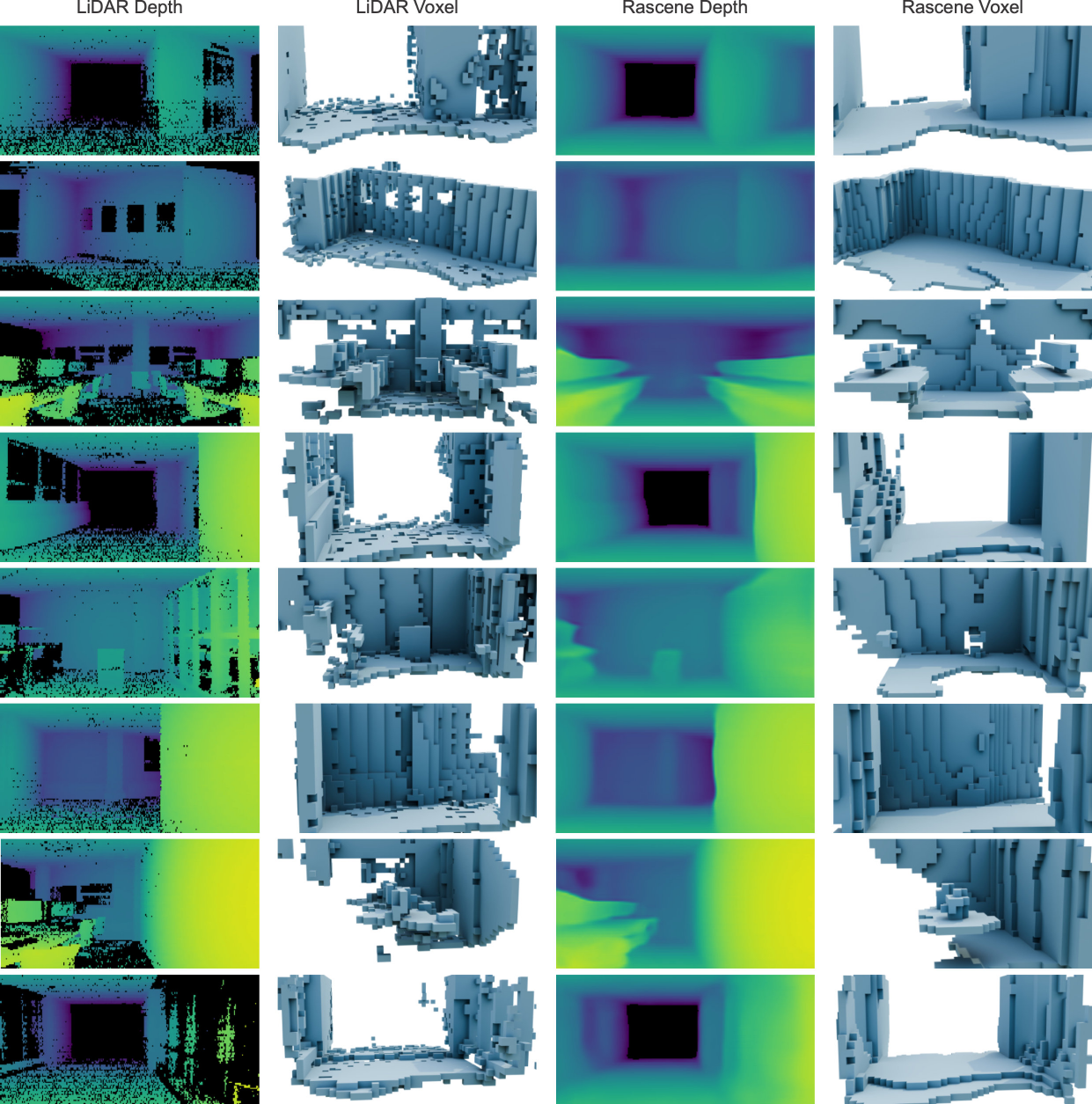}
    \caption{Qualitative results of our \pname system. 
    For each row, we show the ground truth depth map and voxel grid derived from the LiDAR sensor (Cols 1-2), 
    and our corresponding predictions generated using mmWave communication signal (Cols 3-4).}\vspace{-0.05in}
    \label{fig:qualitative-performance}\vspace{-0.3in}
    \end{minipage}
\end{figure*}

\section{Experiments}

\subsection{Main Results}

\begin{table}
\footnotesize
\centering
\caption{Within-dataset comparison with baselines.}\vspace{-0.05in}
\resizebox{\linewidth}{!}{
\begin{tabular}{*{6}{c}}
\toprule
Methods & \# of frame & AbsRel & MAE & CD & CD$_{\text{Diag}}$ \\
\midrule
PanoRadar \cite{lai2024enabling} & 1 & 14.7\% & 34.1 & 32.2 & 3.8\% \\
CartoRadar \cite{cartoradar} & 5 & --- & --- & 26.8 & 3.1\% \\
\pname(Ours) & 1 & 14.1\% & 32.9 & 31.6 & 3.6\% \\
\textbf{\pname(Ours)} & \textbf{5} & \textbf{9.4\%} & \textbf{20.2} & \textbf{19.7} & \textbf{2.3\%} \\
\bottomrule
\end{tabular}
}\vspace{-0.1in}
\label{table:within-dataset-comparison}
\end{table}

\noindent\textbf{Quantitative Results.}
We first evaluate cross-scene generalization by training on 12 environments and testing on 8 unseen environments (A--H), with results in Tab.~\ref{table:quantitative-performance}.
Across all test scenes, \pname achieves strong average performance on both depth estimation and voxel reconstruction.
Although error varies with scene difficulty (\textit{e.g.}, ``E''/``F'' are harder than ``C''/``D''), the normalized Chamfer Distance CD$_{\text{Diag}}$ remains consistently low (1.1\%--3.7\%), indicating stable geometric quality across environments.

\vspace{0.02in}
\noindent\textbf{Qualitative Results.}
Fig.~\ref{fig:qualitative-performance} provides qualitative comparisons between LiDAR targets (Cols 1--2) and \pname predictions from mmWave communication signals (Cols 3--4).
In many scenes, LiDAR targets contain no-return regions (black pixels), commonly caused by absorption on low-albedo surfaces (\textit{e.g.}, dark carpets) and specular reflection on smooth materials (\textit{e.g.}, glass).
Despite these challenging regions, \pname recovers coherent scene geometry, highlighting the complementary robustness of RF sensing to optical material failure modes.

\vspace{0.02in}
\noindent\textbf{Comparison with Baselines.}
For a fair comparison, we implement the models of PanoRadar \cite{lai2024enabling} and CartoRadar \cite{cartoradar} and evaluate all methods on the same within-dataset split.
As shown in Tab.~\ref{table:within-dataset-comparison}, \pname with 5-frame fusion achieves the best overall performance.
More importantly, compared with CartoRadar, our model benefits more from multi-frame, validating the effectiveness of our multi-frame RF fusion module.

\subsection{Ablation Study}

\begin{figure*}[!t]
    \centering
    \includegraphics[width=\textwidth]{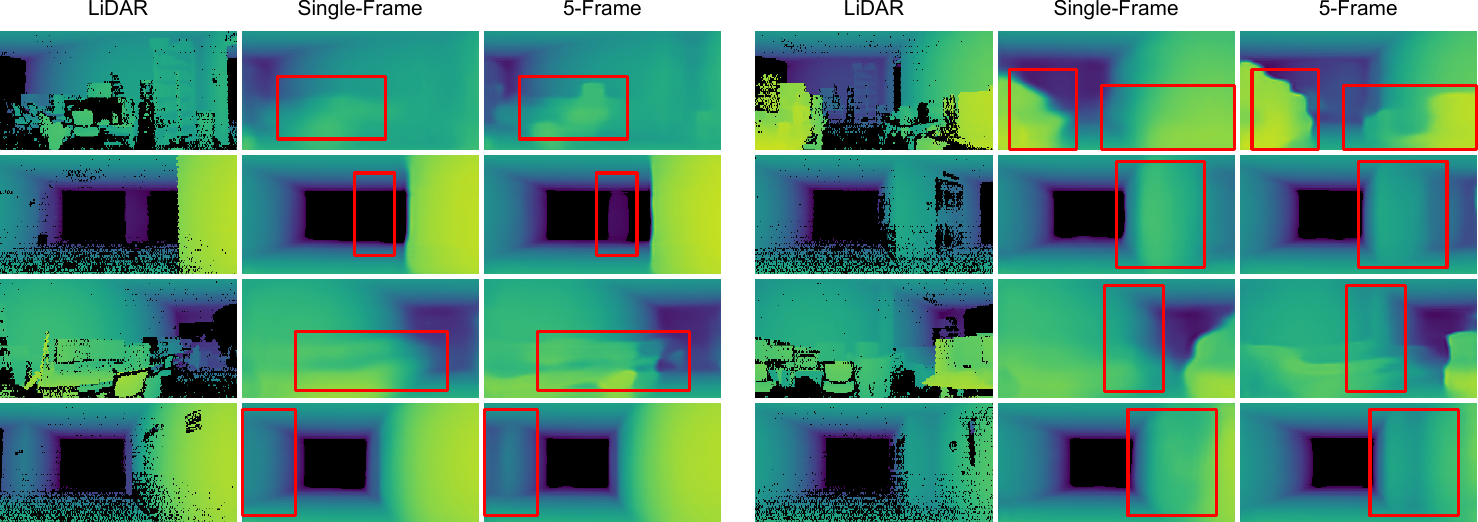}\vspace{-0.05in}
    \caption{Qualitative comparison of single-frame and 5-frame predictions. 
    Multi-frame fusion reduces missed detections and suppresses hallucinated structures, yielding more complete and geometrically consistent reconstructions closer to the LiDAR ground truth.}\vspace{-0.1in}
    \label{fig:num_frame}
\end{figure*}

\begin{table}
\footnotesize
\centering
\caption{Impact of the number of fused radio frames ($N$ in Eq.~\eqref{eq:fusion-mechanism}).}\vspace{-0.05in}
\begin{tabular}{
m{0.13\linewidth}<{\centering}
m{0.11\linewidth}<{\centering}
m{0.11\linewidth}<{\centering}
m{0.11\linewidth}<{\centering}
m{0.11\linewidth}<{\centering}
m{0.11\linewidth}<{\centering}}
\toprule
\multirow{2}{*}{\vspace{-0.07in} Frame \#} & \multicolumn{3}{c}{Depth} & \multicolumn{2}{c}{Voxel} \\ 
\cmidrule(lr){2-4}\cmidrule(lr){5-6}
& AbsRel & MAE & RMSE & CD & CD$_\text{Diag}$ \\
\midrule
1 & 14.1\% & 32.9 & 56.2 & 31.6 & 3.6\% \\
2 & 11.1\% & 24.6 & 44.2 & 26.0 & 3.0\% \\
3 & 9.8\% & 21.8 & 40.4 & 21.9 & 2.5\% \\
5 & 9.4\% & 20.2 & 38.0 & 19.7 & 2.3\% \\
\bottomrule
\end{tabular}
\label{table:ablation-multi-frame}
\end{table}

\begin{table}
\footnotesize
\centering
\caption{Impact of pose inaccuracies (rotation and translation perturbations added at test time).}\vspace{-0.05in}
\begin{tabular}{
m{0.06\linewidth}<{\centering}
m{0.17\linewidth}<{\centering}
m{0.10\linewidth}<{\centering}
m{0.07\linewidth}<{\centering}
m{0.08\linewidth}<{\centering}
m{0.06\linewidth}<{\centering}
m{0.09\linewidth}<{\centering}}
\toprule
\multicolumn{2}{c}{\multirow{2}{*}{\vspace{-0.07in} Error Range}} & \multicolumn{3}{c}{Depth} & \multicolumn{2}{c}{Voxel} \\ 
\cmidrule(lr){3-5}\cmidrule(lr){6-7}
& & AbsRel & MAE & RMSE & CD & CD$_\text{Diag}$ \\
\midrule
\multirow{3}{*}{\vspace{-0.01in} Rot.} & $0^\circ \sim 5^\circ$ & 11.7\% & 25.8 & 46.8 & 23.1 & 2.6\% \\
& $5^\circ \sim 10^\circ$ & 18.4\% & 40.0 & 66.8 & 31.2 & 3.6\% \\
& $10^\circ \sim 15^\circ$ & 18.8\% & 42.0 & 67.7 & 34.3 & 3.9\% \\
\midrule
\multirow{3}{*}{\vspace{-0.01in} Trans.} & $0 \sim 5$ cm & 9.4\% & 20.2 & 38.1 & 19.7 & 2.3\% \\
& $5 \sim 10$ cm & 9.4\% & 20.4 & 38.2 & 19.8 & 2.3\% \\
& $10 \sim 15$ cm & 9.5\% & 20.6 & 38.5 & 20.0 & 2.3\% \\
\midrule
\multicolumn{2}{c}{No Perturbation} & 9.4\% & 20.2 & 38.0 & 19.7 & 2.3\% \\
\bottomrule
\end{tabular}\vspace{-0.1in}
\label{table:ablation-pose-error}
\end{table}

\noindent\textbf{Impact of Multi-Frame Fusion.}
We study the effect of the number of fused frames \(N\) in Eq.~\eqref{eq:fusion-mechanism}.
Tab.~\ref{table:ablation-multi-frame} shows that increasing \(N\) from 1 to 5 steadily improves both depth and voxel metrics.
The most significant gain occurs when moving from 1 to 2 frames, indicating that even one additional viewpoint introduces strong geometric constraints for disambiguation.
Fig.~\ref{fig:num_frame} further shows that multi-frame fusion suppresses hallucinated structures and alleviates missed detections, resulting in more complete geometry.

\vspace{0.02in}
\noindent\textbf{Robustness to Pose Inaccuracies.}
We evaluate robustness to pose noise by perturbing ground-truth poses at test time.
As shown in Tab.~\ref{table:ablation-pose-error}, our model remains highly stable under translation perturbations while being more sensitive to rotation errors.
This trend is geometrically expected, since angular errors are amplified with increasing range.
For example, a rotation error of 5$^\circ$ induces an offset of 61.1\,cm at a distance of 7\,m.

\begin{figure}
    \centering
    \includegraphics[width=\linewidth]{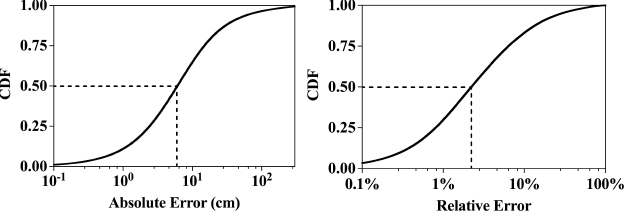}\vspace{-0.05in}
    \caption{Cumulative distribution functions illustrating absolute and relative depth estimation errors.}
    \label{fig:performance-cdf}
\end{figure}

\begin{figure}
    \centering
    \includegraphics[width=\linewidth]{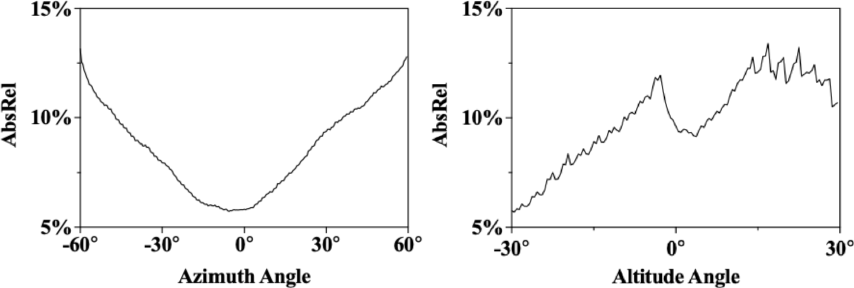}\vspace{-0.05in}
    \caption{Evaluation of depth estimation accuracy across the azimuth and altitude dimensions of the sensor field of view.}
    \label{fig:angle-error}
\end{figure}

\begin{figure}
    \centering
    \includegraphics[width=\linewidth]{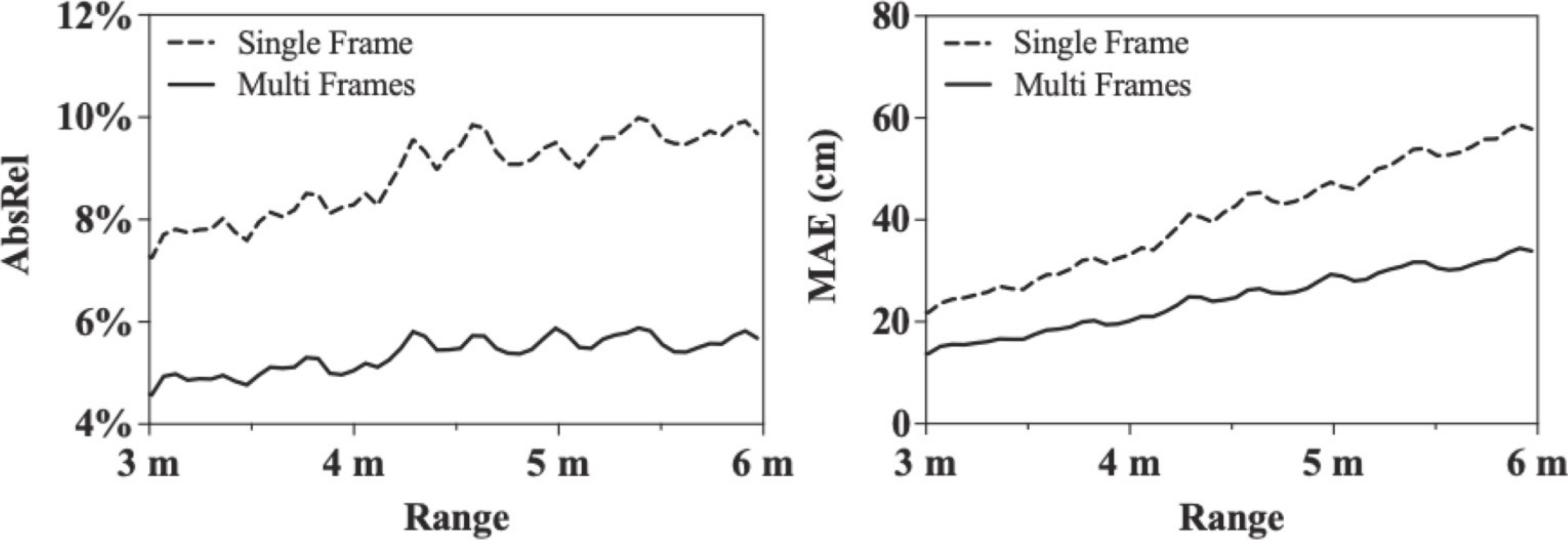}\vspace{-0.05in}
    \caption{Depth error as a function of range for single-frame and multi-frame inference.}\vspace{-0.1in}
    \label{fig:range-error}
\end{figure}

\vspace{0.02in}
\noindent\textbf{Cumulative Error Analysis.}
Fig.~\ref{fig:performance-cdf} reports the cumulative distributions of per-pixel absolute and relative depth errors.
The median absolute error is 6.1\,cm, and 90\% of pixels are below 37.6\,cm; the median relative error is 2.2\%, with a 90th-percentile value of 16.2\%.
This long-tail distribution indicates that most pixels are reconstructed accurately, 
while the overall mean error is affected by a small fraction of large-error outliers, 
which typically located near depth discontinuities or in regions with severe signal attenuation.

\vspace{0.02in}
\noindent\textbf{Field of View Error Analysis.}
We further analyze how depth accuracy varies across the sensor field of view.
Fig.~\ref{fig:angle-error} shows the AbsRel across azimuth and altitude bins.

Along the azimuth axis, the error follows a U-shape: it is lowest near boresight (\(\sim\)5.9\% at \(0^\circ\)) and increases toward the boundaries (\(\sim\)13\% at \(\pm 60^\circ\)), consistent with reduced beamforming gain at the edges.
Along the altitude axis, the error is lower near the floor (\(\sim\)5.7\% near \(-30^\circ\)), peaks around mid-altitude (\(\sim\)12\%), and then slightly decreases toward the ceiling, suggesting that scene complexity is the dominant factor along the vertical direction.

The mid-altitude region typically contains cluttered objects, which are inherently more difficult to reconstruct. 
In contrast, the floor and ceiling are dominated by large planar structures, which are easier to reconstruct due to their geometric simplicity, even under lower antenna gain.

\vspace{0.02in}
\noindent\textbf{Range-Dependent Error Analysis.}
We analyze depth error as a function of range for single-frame and multi-frame inference in Fig.~\ref{fig:range-error}.
While the error increases with distance due to signal attenuation, our multi-frame RF fusion method substantially alleviates this trend.
From 3 to 6\,m, the absolute relative error increase is 2.4\% for single-frame inference but only 1.1\% for multi-frame inference.
Over the same range, multi-frame fusion reduces mean absolute error from 41\,cm to 24\,cm, demonstrating long-range robustness.

\begin{table}
\footnotesize
\centering
\caption{Evaluation of \pname's robustness to common occluders.}\vspace{-0.05in}
\begin{tabular}{
m{0.18\linewidth}<{\centering}
m{0.10\linewidth}<{\centering}
m{0.10\linewidth}<{\centering}
m{0.10\linewidth}<{\centering}
m{0.10\linewidth}<{\centering}
m{0.10\linewidth}<{\centering}}
\toprule
\multirow{2}{*}{\vspace{-0.07in} Occlusion} &
\multicolumn{3}{c}{Depth} & \multicolumn{2}{c}{Voxel} \\ 
\cmidrule(lr){2-4}\cmidrule(lr){5-6}
& AbsRel & MAE & RMSE & CD & CD$_\text{Diag}$ \\
\midrule
No Occ. & 5.6\% & 13.7 & 24.8 & 18.4 & 2.0\% \\
Paper Sheet & 6.1\% & 14.4 & 25.2 & 19.1 & 2.1\% \\
Styrofoam & 6.4\% & 15.7 & 27.4 & 21.2 & 2.3\% \\
\bottomrule
\end{tabular}
\label{table:ablation-occlusion}
\end{table}

\begin{figure}
    \centering
    \includegraphics[width=\linewidth]{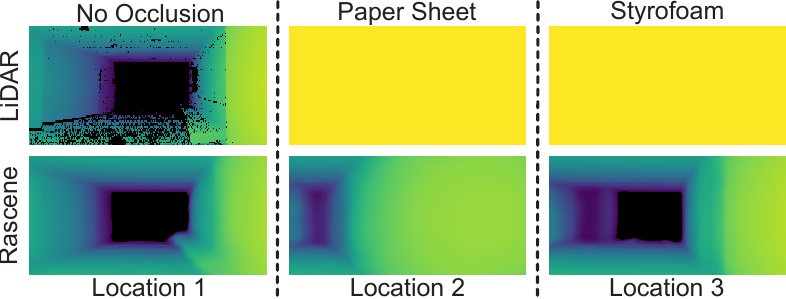}\vspace{-0.05in}
    \caption{Representative examples of different sensors' occlusion resilience in a corridor scene at different locations.}\vspace{-0.10in}
    \label{fig:performance-occ}
\end{figure}

\vspace{0.02in}
\noindent\textbf{Robustness to Occlusion.}
Finally, we evaluate \pname's robustness to occlusion, a key advantage of RF sensing over optical sensors (Tab.~\ref{tab:isac_comparison}).
We place two common occluders, a paper sheet and a styrofoam box, in front of the device.
As shown in Tab.~\ref{table:ablation-occlusion}, the performance degradation is minimal, indicating that \pname can effectively sense through these materials.
Fig.~\ref{fig:performance-occ} further provides a qualitative comparison. 
While LiDAR is completely blocked by occluders, \pname still reconstructs the underlying scene geometry.

\section{Conclusion}

In this paper, we presented \pname, a monostatic ISAC framework that enables high-fidelity 3D scene imaging on individual mmWave communication devices. 
By leveraging CIR measurements from full-duplex OFDM communication devices and a confidence-aware multi-frame fusion strategy across arbitrary poses, \pname mitigates the sparsity, noise, and multipath ambiguity of single-frame RF observations. 
Experiments across diverse indoor environments demonstrate strong cross-scene generalization, consistent gains from multi-frame fusion, and robustness to common occlusions.
These results highlight the potential of using commodity communication infrastructure for low-cost, scalable, and robust 3D perception, especially in scenarios where optical sensing is unreliable.

\clearpage
{
    \small
    \bibliographystyle{ieeenat_fullname}
    \bibliography{reference}
}

\clearpage
\setcounter{page}{1}
\maketitlesupplementary

\section{Monostatic ISAC Hardware}


\begin{figure}
    \centering
    \includegraphics[width=\linewidth]{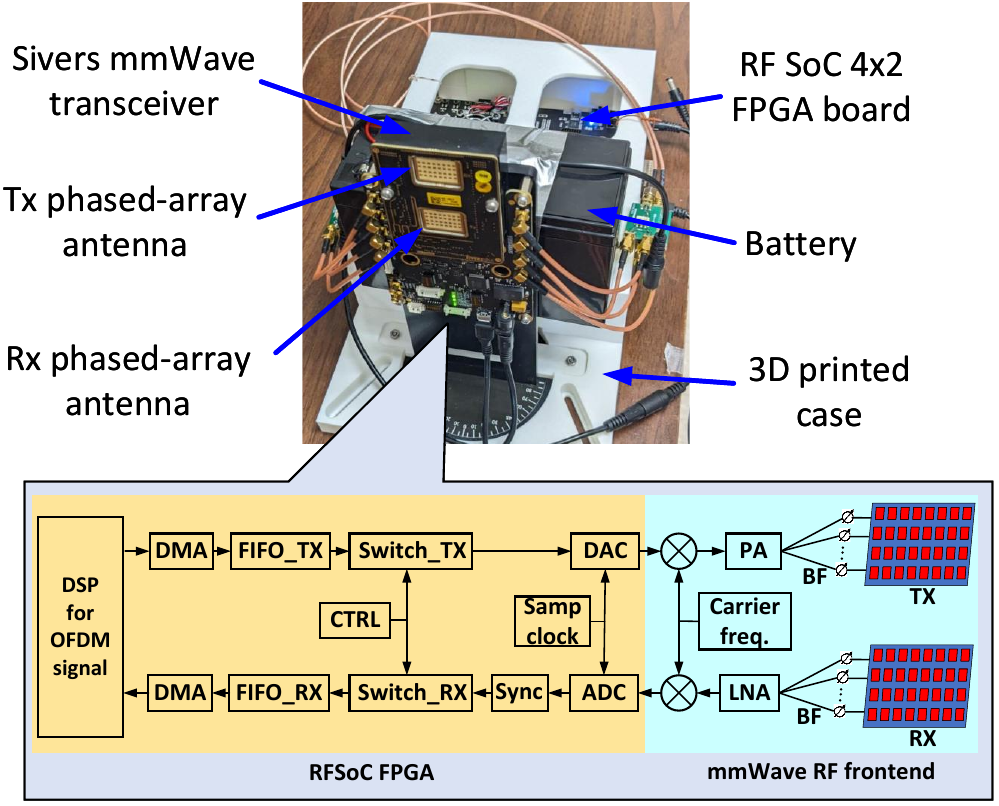}\vspace{-0.05in}
    \caption{Our prototyped monostatic ISAC device.}\vspace{-0.1in}
    \label{fig:isac}
\end{figure}

We built a monostatic ISAC prototype using commercial off-the-shelf (COTS) components, enabling joint communication and sensing within a compact device.

Fig.~\ref{fig:isac} shows our monostatic ISAC prototype, with its parameters summarized in Tab.~\ref{tab:system_params}. The system consists of two primary COTS modules:
(i) an AMD/Xilinx RFSoC 4x2 FPGA board, and
(ii) a Sivers mmWave transceiver with Tx/Rx phased-array antennas.
The RFSoC FPGA implements an OFDM signal processing pipeline compatible with 5G and Wi-Fi protocols, while the mmWave transceiver handles 60\,GHz radio transmission and reception. Within the FPGA, the transmission and reception pipelines are jointly optimized and precisely calibrated to ensure timing and phase alignment required for monostatic sensing. Phased-array antenna control is seamlessly integrated into the processing pipeline, enabling beam steering to be synchronized with signal transmission.

\begin{figure}[h]
    \centering
    \includegraphics[width=\linewidth]{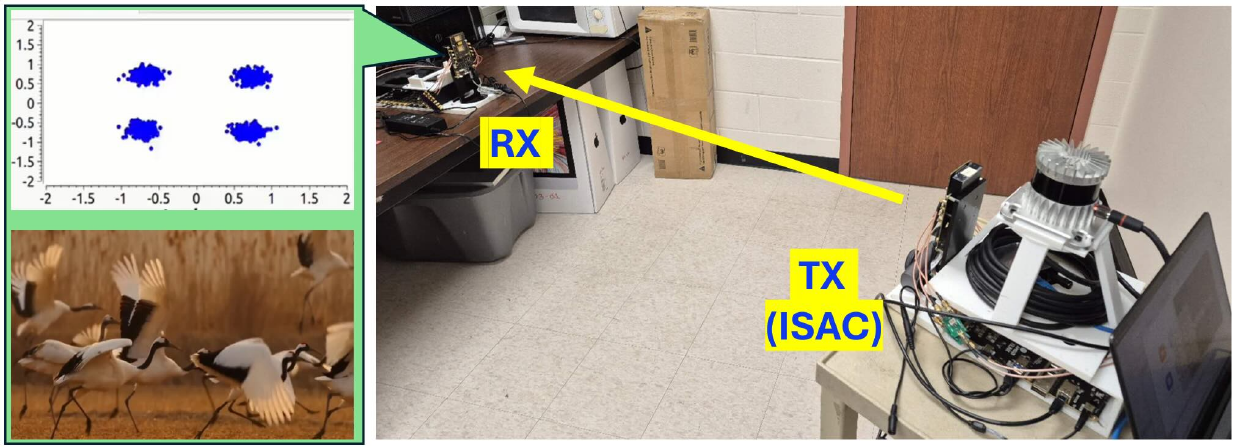}
    \caption{Illustration of video streaming data communication during sensing data collection.}
    \label{fig:isac_a}
\end{figure}

Fig.~\ref{fig:isac_a} shows an example of simultaneous sensing and communication using the prototyped \pname. During sensing data collection, \pname simultaneously sends data packets to another device, supporting continuous video streaming. Both sensing and communication share the same hardware, spectrum band, and radiated energy.
Our implementation demonstrates that a monostatic ISAC device can be realized without specialized sensing hardware. Furthermore, because the design leverages COTS communication components, it can be readily integrated into existing 5G and Wi-Fi mmWave communication devices through firmware and software upgrades.

\begin{table}[t]
\centering
\footnotesize
\caption{The parameters of our monostatic ISAC device.}\vspace{-0.05in}
\renewcommand{\arraystretch}{1.2}
\begin{tabular}{|ll|}
\hline
\textbf{Hardware parameters} & \\
\hline
Sampling rate & 1.2288~GSPS \\
Number of Tx antennas & 16 \\
Number of Rx antennas & 16 \\
Center frequency & 60~GHz \\
Transmission power & 20~dBm \\
\hline
\textbf{Communication parameters} & \\
\hline
Waveform & OFDM \\
FFT points & 1024 \\
Number of valid subcarriers & 900 \\
Cyclic prefix length & 276 \\
OFDM symbol duration & 1.057~$\mu$s \\
Number of OFDM symbols per frame & $16\times16 = 256$ \\
Supporting protocols & 5G and Wi-Fi \\
\hline
\textbf{Sensing parameters} & \\
\hline
Detection time of a frame & 67~$\mu$s \\
Number of frames per second & 10 \\
Theoretical detection range & 30~m \\
Practical detection range & 10~m \\
Number of effective antennas (horizontal) & 8 \\
Number of effective antennas (elevation) & 4 \\
Horizontal antenna spacing & 0.5~wavelength \\
Elevation antenna spacing & 0.5~wavelength \\
Horizontal field of view & [-60$^\circ$, 60$^\circ$]\\
Elevation field of view & [-30$^\circ$, 30$^\circ$]\\
\hline
\end{tabular}
\label{tab:system_params}
\end{table}

\begin{figure*}
    \centering
    \includegraphics[width=\linewidth]{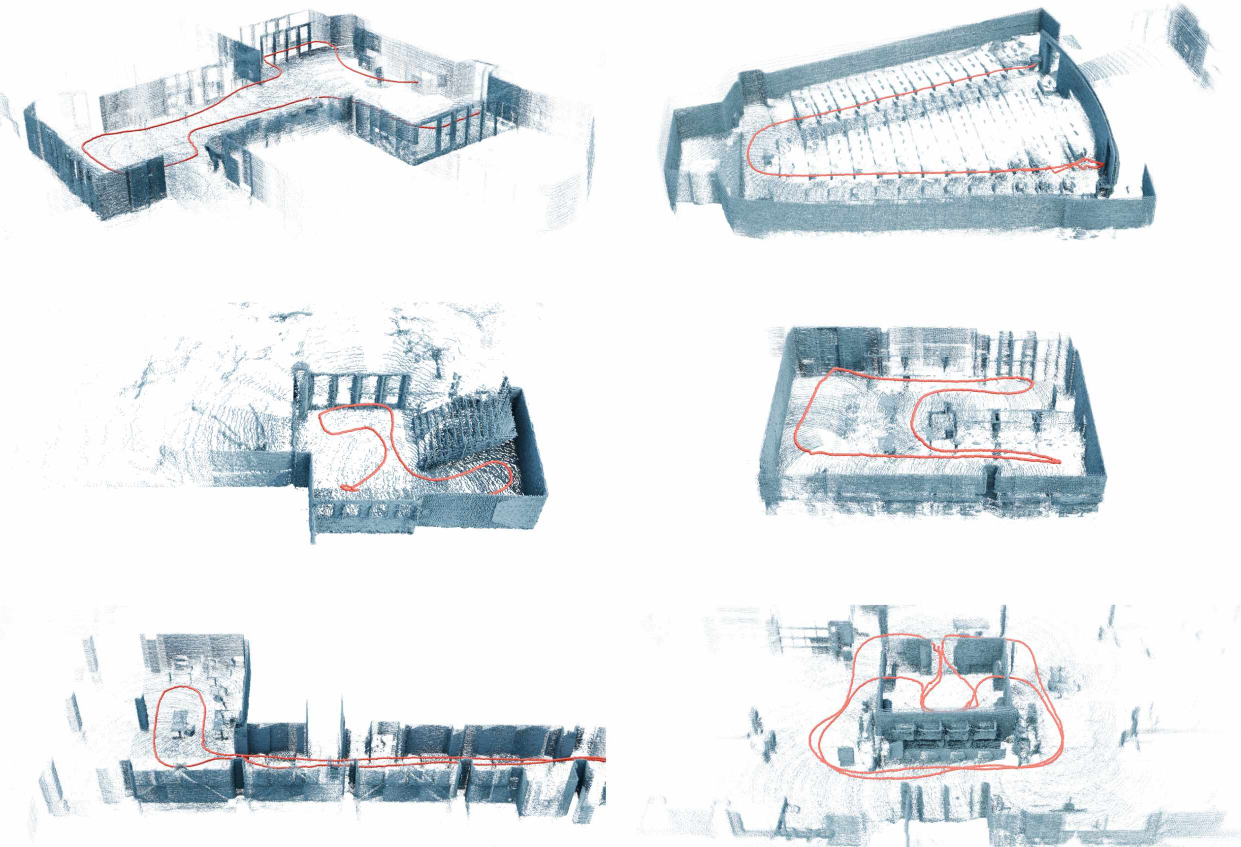}
    \caption{Sample trajectory segments (shown in red) from different scenes, visualized on the ground truth LiDAR point clouds.}\vspace{-0.1in}
    \label{fig:collect-traj}
\end{figure*}

\section{Data Collection}

\begin{figure}
    \centering
    \includegraphics[width=\linewidth]{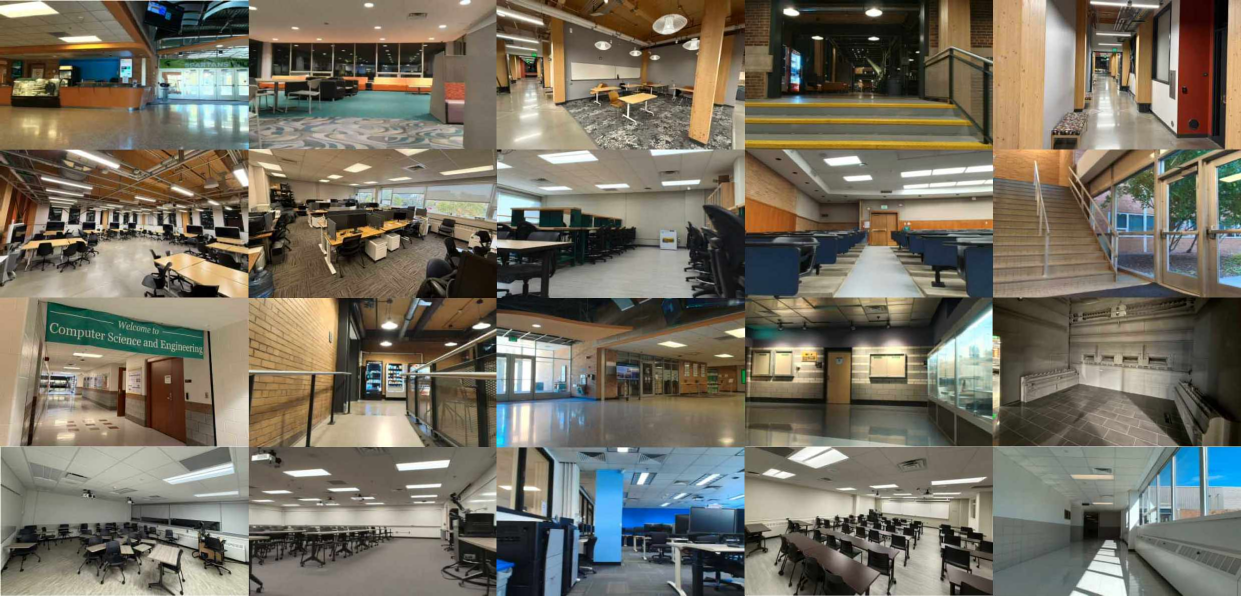}
    \caption{Example snapshots from the 20 distinct indoor environments included in our dataset.}\vspace{-0.1in}
    \label{fig:collect-scene}
\end{figure}

\noindent\textbf{Platform.}
To collect paired RF-LiDAR data, we mounted our custom-designed ISAC device, an Ouster OS0-128 LiDAR, and a TDK ICM-20948 IMU on a movable cart. 
The final dataset contains synchronized RF-LiDAR frame pairs collected from 20 indoor environments spanning diverse layouts, clutter levels, and construction materials such as drywall, glass, and metal. 
For each environment, we recorded approximately 10-20 minutes of data while manually moving the cart along unconstrained trajectories.
Example trajectory segments and scene snapshots are shown in Fig.~\ref{fig:collect-traj} and Fig.~\ref{fig:collect-scene}, respectively.
All sensing modalities, including radio frames, LiDAR scans, and IMU measurements, were synchronized using timestamps from a shared clock source.

\vspace{0.02in}
\noindent\textbf{Ground Truth.}
The Ouster OS0-128 LiDAR provides a 360$^\circ$ horizontal and 90$^\circ$ vertical field of view, whereas our ISAC sensor covers 120$^\circ$ horizontally and 60$^\circ$ vertically.
To align the two modalities, we calibrated the fixed extrinsic transformation between the rigidly mounted sensors and used it to crop the panoramic LiDAR observations to the field of view of the ISAC sensor.
The cropped high-resolution LiDAR point clouds are then used to derive the ground-truth 3D geometry $\mathbf{V}^*$ and depth maps $\mathbf{D}^*$ for training and evaluation.

\vspace{0.02in}
\noindent\textbf{Temporal Sampling Strategy.}
Both the LiDAR and ISAC streams are recorded at 10 Hz, while the platform moves at an average speed of 0.5 m/s. 
Directly using consecutive frames would yield only a small spatial baseline and limited parallax, making it difficult to disambiguate true scene structure from multipath artifacts. 
We therefore adopt a sparse temporal sampling strategy: one frame is selected every 2\,s from the continuous streams, and five sampled frames are grouped into one input window. 
This design preserves sufficient spatial overlap for cross-frame geometric consensus while introducing enough viewpoint variation to provide useful parallax and more diverse multipath observations. 
After warping the frames into a shared reference coordinate system, true scene structures remain more consistent across views than multipath artifacts, which makes the fusion process more reliable.

\end{document}